\documentclass[10pt,twocolumn,letterpaper]{article}

\usepackage{cvpr}
\usepackage{times}
\usepackage{epsfig}
\usepackage{graphicx}
\usepackage{amsmath}
\usepackage{amssymb}
\usepackage{textcomp}
\usepackage[export]{adjustbox}
\usepackage{ctable}
\usepackage{colortbl}
\usepackage[tight,footnotesize]{subfigure}

\usepackage[pagebackref=true,breaklinks=true,letterpaper=true,colorlinks,bookmarks=false]{hyperref}
\newcommand{\name}[0]{CT-Net }

 \cvprfinalcopy 
\usepackage{authblk}

\def\cvprPaperID{3329} 

\ifcvprfinal\fi
\begin{document}

\title{Lose The Views: Limited Angle CT Reconstruction via Implicit Sinogram Completion}

\author[]{Rushil Anirudh\thanks{Corresponding author: anirudh1@llnl.gov}}
\author[]{Hyojin Kim}
\author[]{Jayaraman J. Thiagarajan}
\author[]{K. Aditya Mohan}
\author[]{Kyle Champley}
\author[]{Timo Bremer}
\affil[]{Lawrence Livermore National Laboratory}

\maketitle

\begin{abstract}

Computed Tomography (CT) reconstruction is a fundamental component to a wide variety of applications ranging from security, to healthcare. The classical techniques require measuring projections, called sinograms, from a full 180\textdegree~view of the object. However, obtaining a full-view is not always feasible, such as when scanning irregular objects that limit flexibility of scanner rotation. The resulting limited angle sinograms are known to produce highly artifact-laden reconstructions with existing techniques. In this paper, we propose to address this problem using CTNet -- a system of 1D and 2D convolutional neural networks, that operates directly on a limited angle sinogram to predict the reconstruction. We use the x-ray transform on this prediction to obtain a ``completed'' sinogram, as if it came from a full 180\textdegree view. We feed this to standard analytical and iterative reconstruction techniques to obtain the final reconstruction. We show with extensive experimentation on a challenging real world dataset that this combined strategy outperforms many competitive baselines. We also propose a measure of confidence for the reconstruction that enables a practitioner to gauge the reliability of a prediction made by  CTNet. We show that this measure is a strong indicator of quality as measured by the PSNR, while not requiring ground truth at test time. Finally, using a segmentation experiment, we show that our reconstruction also preserves the 3D structure of objects better than existing solutions.  
\end{abstract} 
\vspace{-20pt}
\section{Introduction}
Computed Tomography (CT) is one of the most common imaging modalities used in industrial, healthcare, and security settings today. In a typical parallel-beam CT imaging system, x-ray measurements obtained from all viewing angles are effectively combined to produce a cross-sectional image of 3D objects \cite{kak2001principles}. These x-ray measurements are collectively referred to as a \textit{Sinogram}. The inverse problem of reconstructing cross-sectional images (or slices) from raw sinograms has been extensively studied by imaging researchers for several decades (Chapter 3 in \cite{kak2001principles}), the most popular technique being the Filtered Back Projection (FBP), which is derived from a discretization of the closed-form solution for the inverse x-ray transform. Alternatively, iterative techniques such as weighted least squares (WLS) have also been developed that improve upon FBP, in some cases, by making successive approximations of increasing accuracy to obtain the final image. In the traditional CT setting, one assumes access to measurements collected from the full range of views of an object, i.e. $\theta \in [0,180^\circ]$, but increasingly newer techniques are being developed that can recover images when a part of the views are missing, i.e. when $\theta \in [0,\theta_{max}], \theta_{max}<180^\circ$. These are referred to as limited angle projections, and reconstruction in such cases is highly ill-posed, as evidenced by the inferior performance of existing methods. 


\begin{figure}[!t]
	\includegraphics[width=\linewidth]{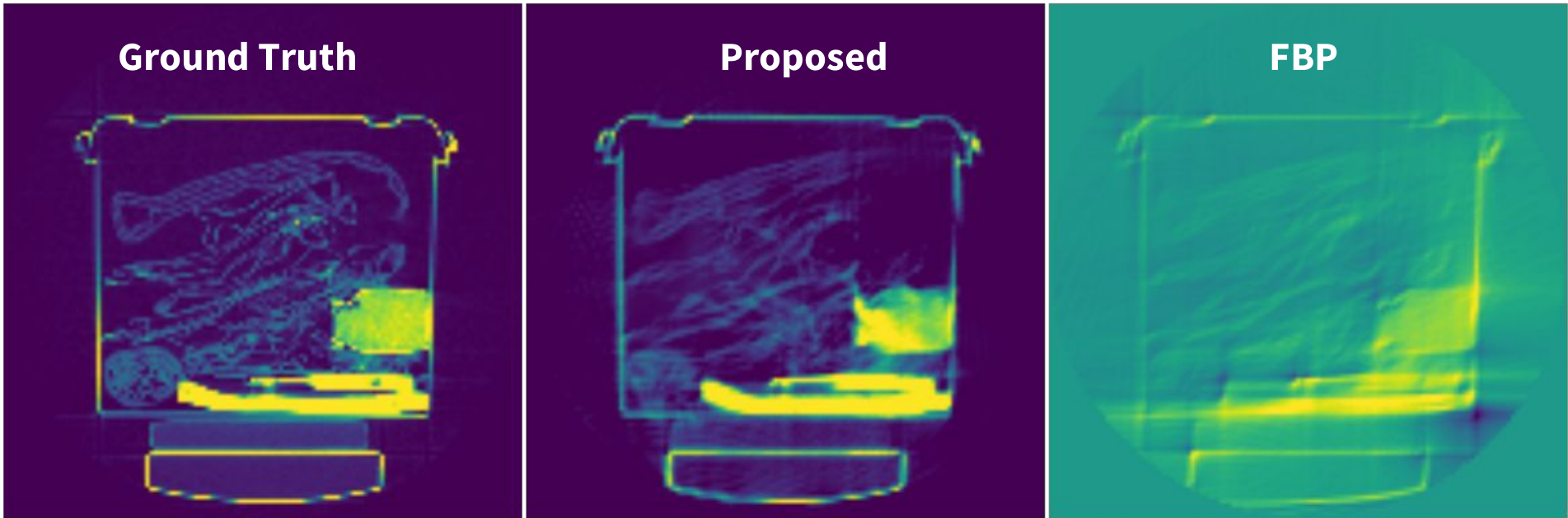}
	\caption{\footnotesize{Proposed limited angle reconstruction compared to FBP and ground truth. These are intensity normalized images}}
	\vspace{-10pt}
	\label{fig:intro}
\end{figure}
\vspace{-10pt}

\paragraph{Need for Limited Angle Scans:} The advantage of such a setup is that it can drastically reduce scan time by restricting the physical movement of the scanner. CT scans are being used to study organs such as the heart, and objects that are highly dynamic, implying that a slightly longer scan time introduces a lot of blurring into the image \cite{mohan2015timbir,cho2013motion}. Further, the limited angle setting can help limit the area of the scan only to a region of interest like in healthcare applications such as breast \cite{niklason1997digital}, and dental \cite{hyvonen2010three} tomography. It can also support applications involving objects that have physical constraints restricting the angles from which they can be scanned, for example in electron microscopy \cite{venkatakrishnan2013model,quinto2008local}. 

\noindent Recent works attempt to solve this problem through a variety of formulations -- as explicit sinogram regression from limited view to full-view \cite{huang2017restoration}, reduction of artifacts obtained from FBP \cite{frikel2013characterization}, and using convolutional neural networks to refine poorly initialized reconstructions obtained from FBP \cite{zhang2016image}. However, these techniques use simpler datasets with very less variability, and operate in regimes where a majority of the viewing angles are captured -- for example $130^\circ,150^\circ$ in \cite{zhang2016image}, $140^\circ$ in \cite{venkatakrishnan2013model} and $170^\circ$ in \cite{huang2017restoration}. Instead, this paper performs CT reconstruction from just half the views during training, i.e. $90^\circ$, from a challenging real world checked-in luggage dataset \cite{coe_data}. 

\vspace{-10pt}
\paragraph{Challenges:} Generally speaking, an edge of an object is recovered accurately in a CT image if an x-ray tangential to that edge is captured. When several such x-rays from a contiguous set of views are missing, as in the limited angle scenario, a significant amount of information regarding the scene is missing. A loose analogy in traditional computer vision, is like reconstructing a scene when it is partially occluded from the camera. In the last few years, data-driven approaches have made significant strides in solving similar challenging image recovery problems \cite{chang2017one} such as image completion \cite{pathak2016context}, image-inpainting \cite{yeh2016semantic}, super-resolution \cite{SRGAN}, CS recovery \cite{Reconnet}. These methods leverage the availability of large datasets, and the expressive power of deep learning to impose implicit constraints to the recovery problem. 

\noindent However, CT reconstruction presents several additional challenges.  Unlike standard images, CT images of transportation luggage, cargo, etc. can be very complex with no apparent low dimensional structure. As a result, even under the classical CT setting of full-view scans, training a neural network end-to-end to predict the final image is challenging. This is further exacerbated by the fact that collecting a large dataset of CT images and their corresponding projections is significantly harder. Consequently, many state-of-the-art methods rely on analytical techniques, e.g. FBP, to provide an initial coarse estimate \cite{huang2017restoration,chen_2017_dl_recon}, which is then refined using deep neural networks. In the extreme setting as considered here, FBP can be highly misleading, rendering subsequent techniques ineffective. An example of FBP under a limited angle setting $(0,90^\circ)$ is shown in Figure \ref{fig:intro}. Finally, CT is typically used in critical applications, which necessitates the need for practitioners to understand the confidence or reliability of reconstructions at test time. 

\vspace{-10pt}
\paragraph{Proposed Work:} In this paper we address these challenges, with \textbf{\name} -- a system consisting of 1D and 2D convolutional neural networks (CNN) coupled with adversarial training to recover CT slices from limited angle sinograms. Since sinograms have certain consistency conditions \cite{willis1995optimal,mazin2010fourier} that are hard to enforce directly within a neural network, we propose to solve this problem by completing sinograms from limited angle to full view (180\textdegree), implicitly in the image space. In other words, we employ a three-stage approach -- first \name produces a reconstruction based on a limited-angle sinogram. Next, we project this image into sinogram space as if it came from full-view measurements using the x-ray transform. Lastly, we use existing techniques such as FBP or WLS to obtain the final image. 
\vspace{5pt}

\noindent We train our network with sinograms containing only half the viewing angles to directly predict ground truth consisting of reconstructions obtained from full-view measurements. Inspired by the success of 1D CNNs in language modeling \cite{kim2014convolutional}, our network interprets the sinogram as a ``sequence''. This formulation allows us to model projections from individual views, while also enabling us to capture relationships across views through a simple attention model. Consequently, our approach supports the use of a different number of views at test time, to even lower viewing angles than $90^\circ$. As seen in Figure \ref{fig:intro}, the proposed sinogram completion strategy is able to recover CT slices with high fidelity much better than FBP. 

\noindent Finally, in order to generate a confidence measure for the recovery process, we propose to estimate per-pixel variabilities to perturbations in the latent space of sinograms, and compute an aggregated confidence score. Interestingly, the proposed score is highly correlated to the actual reconstruction quality measured with respect to the ground truth.


\vspace{5pt}
\noindent Our main contributions can be summarized as follows:
\begin{enumerate}
\item We propose the first deep learning solution to recover CT images from limited angle or incomplete-view sinograms. 
\item We propose to utilize 1D CNNs to process sinograms, which enables generalization to different number of views during training and testing. 
\item We develop a confidence metric for the recovered images, and show that it is a strong indicator of reconstruction quality, as measured by PSNR.
\item We demonstrate that our method significantly outperforms state-of-the-practice approaches on a challenging transportation security dataset.
\item Using 3D semantic segmentation experiments on the resulting reconstructions, we illustrate that the proposed approach preserves the 3D structure effectively.
\end{enumerate}
\section{Other Related Work}
\begin{figure*}[!htb]
\centering
\includegraphics[width=.9\linewidth]{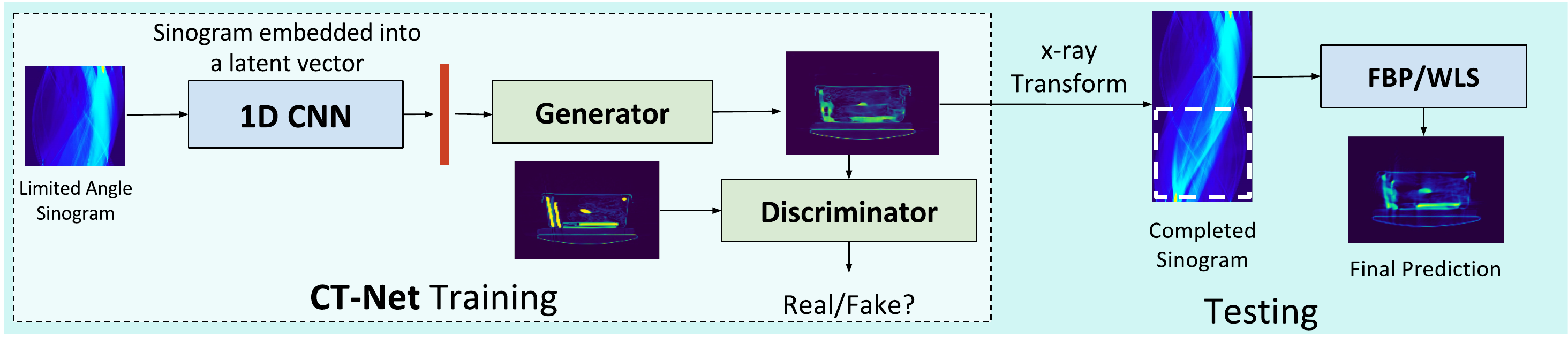}
\caption{\footnotesize{Overview of the proposed approach. During training, we train \name to predict the CT slice directly from limited angle sinograms. We use a mix of mean squared error and an adversarial loss. During test time, we forward project the ouput of \name using the x-ray transform to \textit{complete} the sinogram. Next, we use WLS or FBP on the completed sinogram to obtain the final reconstruction. The architectures for the 1D CNN, Generator and Discriminator are described in the supplementary material.}}
\vspace{-10pt}
\label{fig:overview}
\end{figure*}
There have been other studies addressing limited angle reconstruction in different contexts; we refer the reader to \cite{frikel2013characterization} for a detailed list of them. A related but different problem is the few-view (also called sparse-view) CT reconstruction, which has been of significant interest. It differs from the limited-view problem in that, it reduces the number of viewing angles by \emph{uniformly sampling} in the possible range of angles, $(0^\circ,180^\circ)$ \cite{han2017framing,kang2016deep,jin2017deep}. Zhao \textit{et al.} proposed a convolutional neural network framework to recover poorly reconstructed images \cite{zhao_2016_dl_recon}, and Chen \textit{et al.} used a similar approach to denoise images from low-dose CT \cite{chen_2017_dl_recon}. This recovery process closely resembles the techniques used for solving inverse problems in vision recently such as super-resolution \cite{SRGAN}, recovering images from compressive measurements \cite{Reconnet}, and other linear inverse problems \cite{chang2017one}. When compared to sparse-view reconstruction, the limited angle problem is more challenging, as it is equivalent to extrapolation in the sinogram space. 

\noindent In the tomographic reconstruction community, numerous studies have focused on algebraic approaches to inverse problems, in particular utilizing Algebraic Reconstruction Techniques (ART) and its variants such as Multiplicative ART (MART). Examples include the work on MART-AP \cite{konovalov_2013_fewview} and simultaneous MART \cite{vlasov2015_fewview}. Further, Chen \textit{et al.} proposed an adaptive Non-Local Means (NLM) based reconstruction method to compensate for over-smoothed image edges in few-view reconstructions \cite{chen_2016_fewview}. In addition, there exist methods that utilize dictionary learning techniques, coupled with sparse representations or total variation optimization, for both few-view and low-dose CT reconstruction tasks \cite{lu2012_fewview, xu_2012_dictlearn}. Despite the availability of such varied solutions, to the best of our knowledge, our work is the first that addresses the problem of limited-view CT reconstruction, by directly operating on the limited-view sinogram using viewing angles up to only $90^\circ$. 
\vspace{-10pt}



\section{Preliminaries}
In this section we outline the basics of the CT scan, and briefly describe the state-of-practice algorithms for recovering the tomographic images from the sinograms. 
\subsection{CT Reconstruction problem formulation}

\begin{figure}[!htb]
 \centering
    \subfigure[]{\includegraphics[scale=0.30]{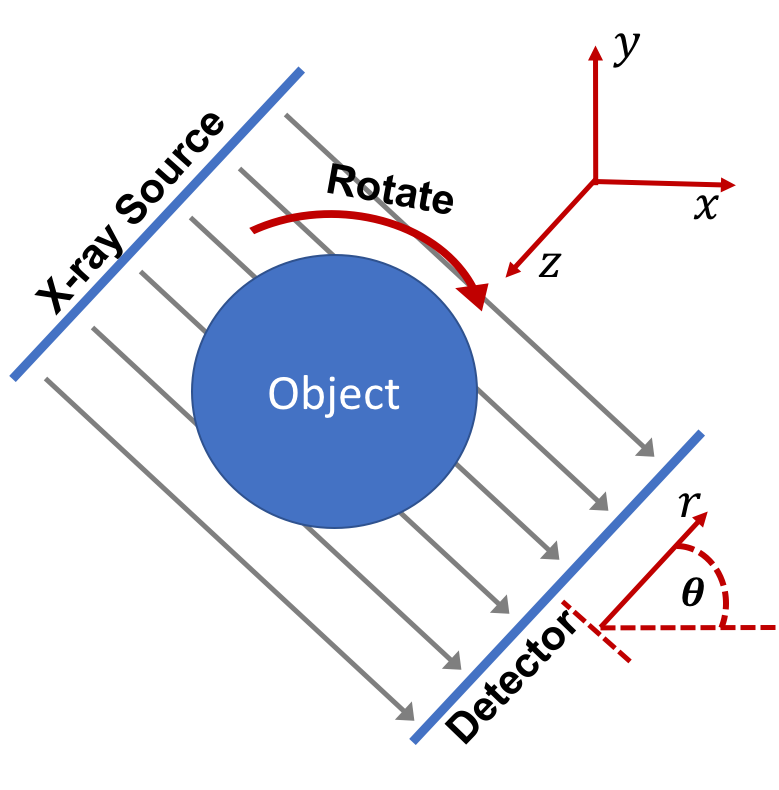}} \hfill
	\subfigure[]{\includegraphics[scale=0.30]{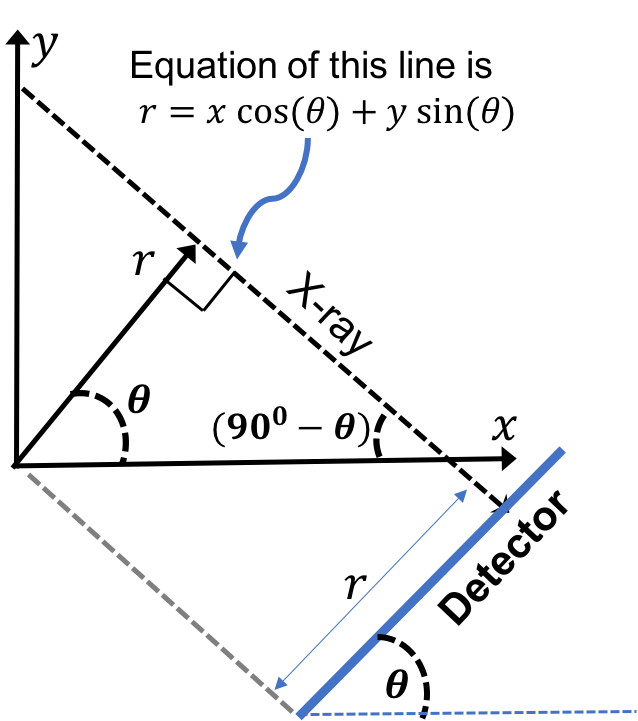}}
\caption{\footnotesize{Experimental setup of x-ray computed tomography. An object is rotated along an axis and exposed to a parallel beam of x-rays.
The intensity of attenuated x-rays exiting the object is measured by the detector at regular angular intervals. 
The projection at an angle of $\theta$ measured at a distance of $r$ on the detector is the line integral of LAC values along the line perpendicular to the detector at $r$. }}
\label{fig:tomosetup}
\vspace{-10pt}
\end{figure}

X-ray CT is a non-destructive imaging modality that is used to reconstruct the interior morphology of an object scanned using x-ray radiation. In our experiments, the object to be imaged is placed in between a source of parallel beam x-rays and a planar detector array. The x-rays get attenuated as they propagate through the object and the intensity of attenuated x-rays exiting the object is measured by the detector. To perform tomographic imaging, the object is rotated along an axis and repeatedly imaged at regular angular intervals of rotation. At each rotation angle of the object, the measurements at the detector can be expressed as the line integration of the linear attenuation coefficient (LAC) values 
along the propagation path. Assume that the object is stationary in the cartesian coordinate system described by the axes $(x,y,z)$.
Then, the projection at a distance of $r$ on the detector is given by,
\begin{equation}
\label{eq:radontrans}
p_{\theta}(r,z) = \int\int \rho(x,y,z) \delta(x\cos(\theta)+y\sin(\theta)-r) dx dy.
\end{equation}
where $\rho(x,y,z)$ is the LAC of the sample at the coordinates $(x,y,z)$, $\delta(~)$ is the standard Dirac delta function, and $\theta$ is the rotation angle. We refer to eq \eqref{eq:radontrans} as the x-ray transform in the remainder of the paper.
Note that rotating the object clockwise by an angle of $\theta$ is equivalent to 
rotating the source and detector pair counterclockwise by an angle of $\theta$ and vice versa.
Notice that the equation \eqref{eq:radontrans} is separable in the $z$ coordinate. Hence, the projection relation is essentially a 2D function in the $x-y$ plane that is repeatedly applied along the $z-$axis. Next, we describe two popular reconstruction algorithms.

\subsection{Filtered Back Projection}
Filtered back-projection (FBP) is an analytic algorithm for reconstructing the sample $\rho(x,y,z)$ from 
the projections $p_{\theta}(r,z)$ at all the rotation angles $\theta$.
FBP directly inverts the relation \eqref{eq:radontrans} to solve for the LAC values $\rho(x,y,z)$.
In FBP, we first compute the Fourier transform $P_{\theta}(\omega)$ of the projection $p_{\theta}(r)$ as a function of $r$. 
The filtered back projection reconstruction $Y$ is then given by \cite{kak2001principles},
\begin{equation}
\label{eq:fbprecon}
Y(x,y) = \int_0^{\pi} Q_{\theta} \left(x\cos(\theta)+y\sin(\theta)\right) d\theta,
\end{equation}where
\begin{equation}
Q_{\theta}(r) = \int_{-\infty}^{\infty} P_{\theta}(\omega) |\omega| \exp(j2\pi\omega r) d\omega.
\end{equation}
From equation \eqref{eq:fbprecon}, we can see that a 
filtered version of $p_{\theta}(r)$ is smeared back on the $x-y$ plane along the direction $(90^\circ-\theta)$ (see figure \ref{fig:tomosetup}). The FBP reconstruction thus consists of the cumulative sum
of the smeared contributions from all the projections ranging from $0^\circ$ to $180^\circ$. 

If the projections are acquired over a limited angular range, 
then the integration in \eqref{eq:fbprecon} will be incomplete in the angular space.
Since each projection $p_{\theta}(r)$ contains the cumulative sum of the LAC
values at a rotation angle of $\theta$, it also contains information about the 
edges that are oriented along the angular direction $(90^\circ-\theta)$ in Figure \ref{fig:tomosetup}. 
Now, suppose data acquisition starts at $\theta=0^\circ$ and stops at an angle of $\theta=\theta^{max}<\pi$.
Then, the edge information contained in the projections at the angles $\theta\in[\theta_{max},\pi]$ will be missing in the final reconstruction.
This is the reason behind the edge blur in the reconstructions shown in figure \ref{fig:intro}.

\subsection{Weighted Least Squares}
Weighted least squares (WLS) is an iterative method for reconstructing the sample
by formulating the reconstruction as the solution to a cost minimization problem. 
It belongs to the class of model-based reconstruction algorithms \cite{wlsweights,mohan2015timbir},
albeit without any form of regularization. 

Let $Y$ represent the reconstruction, that contains all the sampled values of the LAC $\rho(x,y,z)$ in 3D space, and $S$ represent the sinogram. Then, the WLS reconstruction is given by solving the following optimization problem,
\begin{equation}
\label{eq:wlsrecon}
\hat{Y} = \arg\min_Y\left\lbrace (S-AY)^T W (S-AY)\right\rbrace
\end{equation}
where $A$ is the forward projection matrix that implements the line integral of \eqref{eq:radontrans} in discrete space and $W$ is an estimate of  the inverse covariance matrix of $S$ \cite{wlsweights}, computed as a diagonal matrix with $W_{ii} = \exp(-S_i)$.

\section{Proposed Approach}
An overview of the proposed approach is described in Figure \ref{fig:overview}. In this section, we describe the details of our approach, and outline the training and testing strategies. A limited angle sinogram is a collection of measurements of a given object, stored in matrix form, from a range of views spanning from $0^\circ$ to $\theta < 180^\circ$ (fixed at $90^\circ$ in this paper), where each row corresponds to a single view. 
Completing the sinogram directly is challenging, since there are consistency conditions in the sinogram space  \cite{willis1995optimal,mazin2010fourier} that cannot be easily enforced during the training process. Therefore, we resort to an implicit sinogram completion process, that converts a limited-angle sinogram to a full view one as described next.


\subsection{From Half-View to Full-View: Implicit Sinogram Completion with \name}
\label{sec:proposed}
\vspace{5pt}

\noindent \name first embeds the limited angle sinogram into a latent space using a fully convolutional 1D CNN. The 1D convolutions are meaningful in this context, since they allow the use of a simple attention model to study correlations across neighboring views. We interpret the sinograms as a sequence of projections, corresponding to different viewing angles, similar to the sentence modeling in the NLP literature \cite{kim2014convolutional}. In the 1D CNN architecture, we use multiple filters with varying window sizes, in order to capture information across different sized neighborhoods. Each filter produces an embedding corresponding to a window size, resulting in 
the final embedding with dimensions $N_{filters}\times N_{h}$, where $N_h$ denotes the different number of window sizes considered. See \cite{kim2014convolutional} for more details on the implementation. In our case we have $N_{filters}=256$ filters, with $N_h = 5$ window sizes, resulting in a $1280$ dimensional embedding. By design, this formulation supports varying number of rows (views) in the input sinogram. This latent representation is decoded into its corresponding CT image using a 2D CNN to predict the desired CT image. Our decoder is fully convolutional but for a projection layer in the beginning, and consists of residual units \cite{he2016deep}. 
\vspace{5pt}

\noindent \textbf{Training Losses:} We trained \name with the standard mean squared error (MSE) as the loss: $\mathcal{L}_{mse} = ||\hat{Y} - Y||_2^2$, where $\hat{Y}$ and $Y$ denote the predicted and ground truth images respectively. Training with MSE naturally results in the highest PSNR and SSIM metrics, as MSE optimizes specifically for them, however they result in highly smoothed images as the resulting solution is obtained as the average of many possible solutions. 

We also use an adversarial loss \cite{GANGoodfellow}, that uses a discriminator to guide \name to generate more realistic looking reconstructions. In practice, this results in sharper edges and visibly more high frequency content. Similar observations have been reported in the case of super resolution \cite{SRGAN}, where it was found that PSNR is a weak surrogate for visual quality and using adversarial loss produces a more sharper rendering. The adversarial loss is measured as $\mathcal{L}_{adv} = -\log(D(\hat{Y}))$ where $D(.)$ represents the discriminator whose role is to distinguish between a generated image (fake) and an actual slice from the training dataset. The loss at the discriminator is $\mathcal{L}_{D} = -\log(D(Y))-\log(1-D(\hat{Y}))$. The final loss for the generator is hence obtained as $\mathcal{L} = \mathcal{L}_{mse} + \lambda \mathcal{L}_{adv}.$ We found $\lambda = 0.05$ to be a suitable choice, resulting in the best reconstructions. Further details of all the networks inside \name can be found in the supplementary material.

\vspace{5pt}
\noindent \textbf{Sinogram Completion:} Once we have the prediction from CT-Net, $\hat{Y}$, from a $90^\circ$ sinogram, $S_{(0,90)}$, we obtain a full-view sinogram from the $\hat{Y}$ using $\hat{S}_{(0,180)} = \mathcal{F}(\hat{Y},180^\circ)$, where $\mathcal{F}$ corresponds to the x-ray transform (specified in eq \eqref{eq:radontrans}). This computes the sinogram, as if views from $180^\circ$ were available for the current image slice, from which we use rows corresponding to $(90^\circ,180^\circ)$ to obtain the completed sinogram -- $S_{complete} = [S_{(0,90)},\hat{S}_{(90,180)}]$ . We obtain the final reconstruction with $S_{complete}$ using FBP, or WLS using the equations \eqref{eq:fbprecon} or \eqref{eq:wlsrecon} respectively.

\subsection{A New Confidence Score for Reconstructions from Limited Angle Sinograms}
\label{subsec:confidence}
CT reconstruction is often used in critical applications such as healthcare or security, where an incorrect or misleading reconstruction can have negative consequences. This fact is even more important when we perform reconstruction from incomplete data, as we are operating in a highly under-constrained setting. In order to address this, we propose a confidence score, which measures the reliability of the reconstruction, for a given limited angle sinogram. This score is evaluated only at the test time and does not require the ground truth for its estimation.

At test time, the 1D-CNN embeds the limited-angle sinogram into a latent space, where we randomly perturb the vector using a dropout strategy and reconstruct the CT image from the perturbed latent representations. In all our experiments, the dropout probability was set at 0.05. For a given sinogram, we repeat this multiple times, and measure the per-pixel variance in the resulting reconstructions. The intuition here is that if the network has sufficient information to recover the slice, small perturbations in the latent space should not affect the final output significantly. However, if the sinogram does not reliably capture the information in the scene, its corresponding latent representation tends to be unstable, thus leading to significant changes in the reconstruction for small perturbations to its latent vector. Though a simple heuristic such as the $\ell_1$ norm of the per-pixel variances can be directly used as the confidence metric, it can be highly sensitive to the number of objects present in the scan. Hence, we propose the following metric as a confidence score for the reconstruction: For the given per-pixel variance matrix $V$, and the actual predicted image $\hat{Y}_k$ obtained with no latent space perturbation, we define
\begin{equation}
\label{eq:confidence}
r_k = \exp\left(-\frac{\sum_i\sum_j V_{ij}}{||\hat{Y}_k||_2}\right),
\end{equation}
where the variances are normalized by the total $\ell_2$ norm of the reconstructed image. We find that this metric acts as a strong indicator of the actual reconstruction quality, measured as the PSNR with respect to the ground truth image. Since evaluation of this metric does not require ground truth, this can be used by the practitioner to evaluate the reconstruction at test time, without actually generating the ground truth.

\section{Experiments}
\begin{figure}[t]
\centering
	\includegraphics[width=0.85\linewidth]{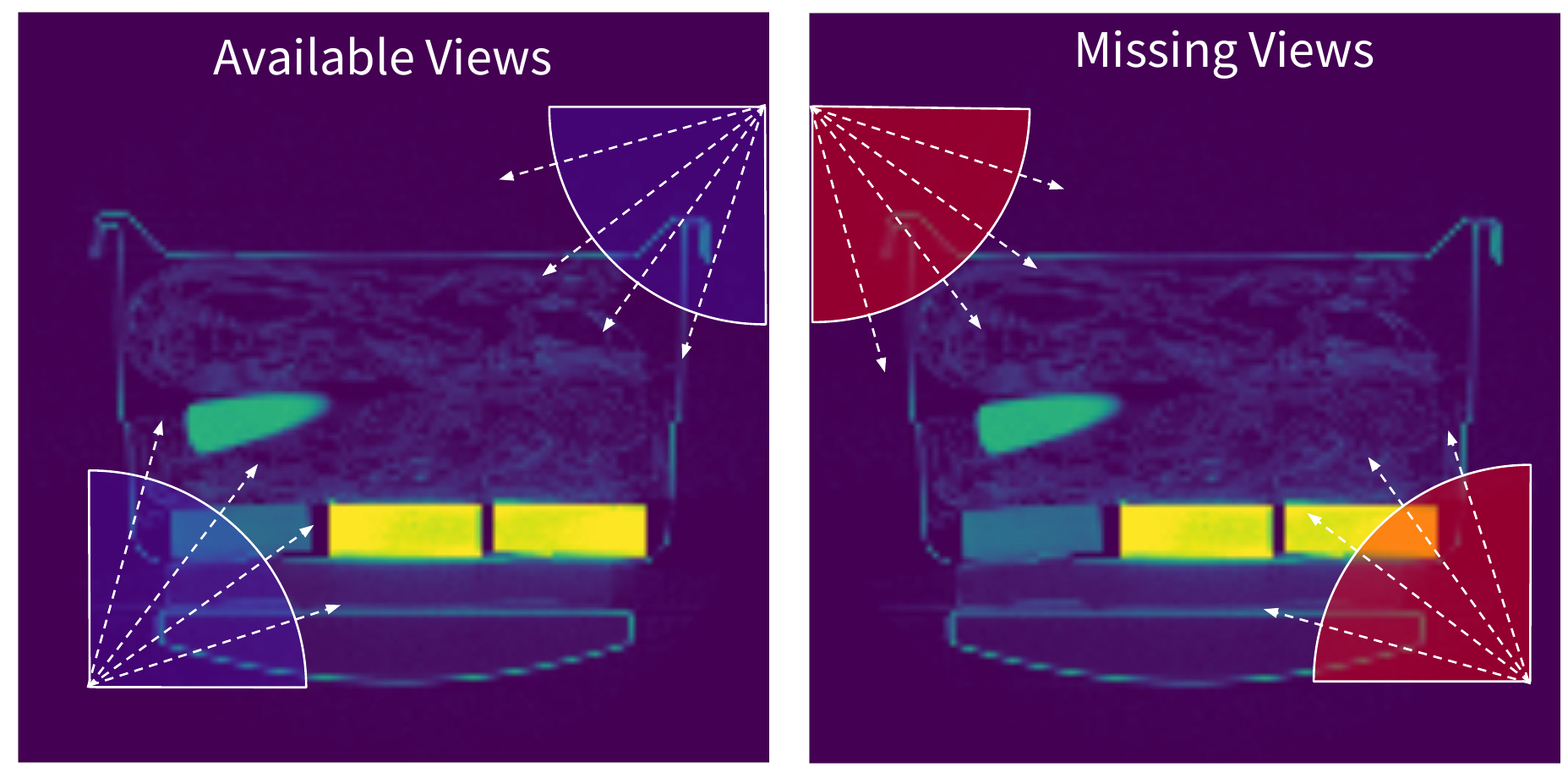}
	\caption{\footnotesize{\textbf{Setup:} The views used in training, shown with reference to the physical arrangement of the training data.}}
	\label{fig:views}
	\vspace{-10pt}
\end{figure}

\begin{figure*}[!htb]
\includegraphics[width=\linewidth]{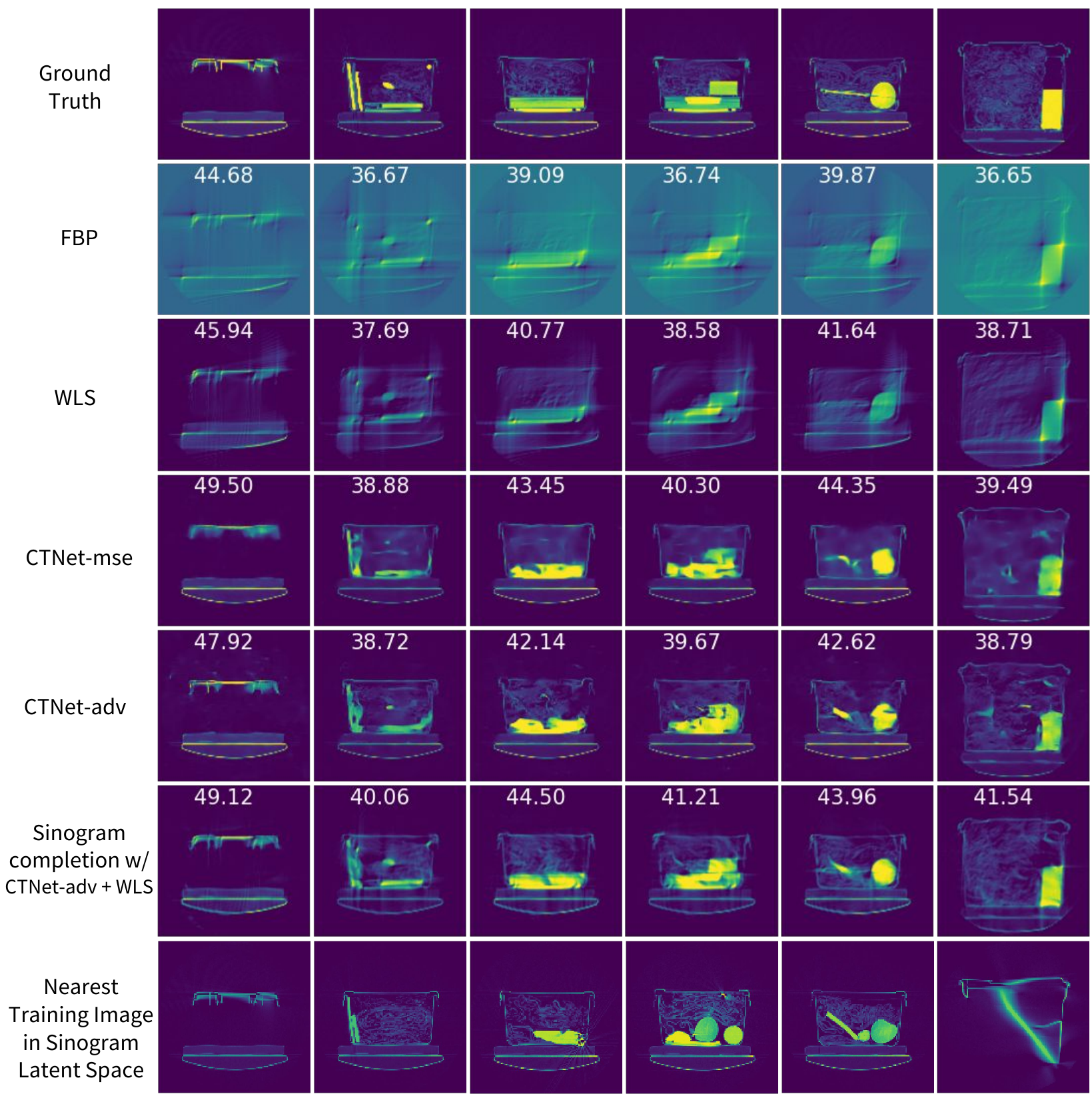}
\caption{\footnotesize{\textbf{Reconstruction Results:} PSNR (dB) is shown for each reconstruction, measured against the ground truth. It should be noted that sinogram completion is much better at preserving geometric shape (such as round objects) than baselines, apart from being superior in terms of PSNR. For sinogram completion, \name-mse and \name-adv performed nearly the same when followed up by WLS. We also show the nearest reconstruction from the training set, based on distance in the sinogram latent space. }}
\label{fig:ctrecon}
\vspace{-10pt}
\end{figure*}
In this section, we compare the sinogram completion approach with several baselines, and demonstrate its effectiveness in industrial CT reconstruction and segmentation. 

\paragraph{Dataset}
We evaluate the our methods on a dataset of CT scans of common checked-in luggage collected using an Imatron electron-beam medical scanner -- a device similar to those found in transportation security systems. The dataset is provided by the DHS ALERT Center of Excellence at Northeastern University \cite{coe_data} for the development and testing of Automatic Threat Recognition (ATR) systems. We repurpose this dataset for generating CT reconstructions from sinograms. The dataset is comprised of $188$ bags, with roughly ~$250$ slices per bag on an average. In total, the dataset consists of ~50K full view sinograms along with their corresponding FBP reconstructions. The original slices are $1024\times 1024$, but we perform experiments on their downsampled versions of size $128\times 128$, and correspondingly the sinograms are subsampled to be of size $720 \times 128$. This corresponds to views obtained at every $0.25$\textdegree sampled uniformly from $180$ \textdegree. We split the bags into a training set of $120$ bags and a test set with the rest. This split resulted in about $35K$ image slices for training and around $15K$ image slices for testing. The bags contain a variety of everyday objects such as clothes, food, electronics etc. that are arranged in random configurations. In all our experiments, we assume access to only the top half of the sinogram, which results in observing half the views $(0^\circ,90^\circ)$ (Figure \ref{fig:views}). We use this partial sinogram to train \name and the different baselines. As an initial processing step, we perform basic filtering to remove low intensity noise in the ground truth reconstructions, obtained using FBP with all the views. 


\begin{figure*}[!htb]
	\centering
	
	\centering
	\subfigure[top row: ground truth, middle: proposed reconstruction, and bottom row: pixel-wise confidence map. Light indicates more confident.]
	{\includegraphics[width=0.32 \linewidth,valign=b]{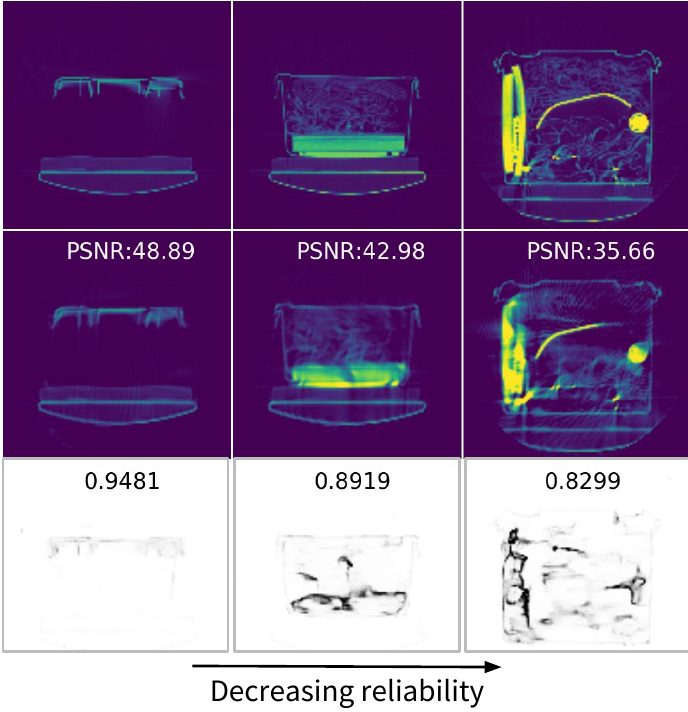}
		\label{fig:sample_confidence}
	}
    \hfill
	\subfigure[PSNR v Reliability]{\includegraphics[width=0.26 \linewidth,valign=b]{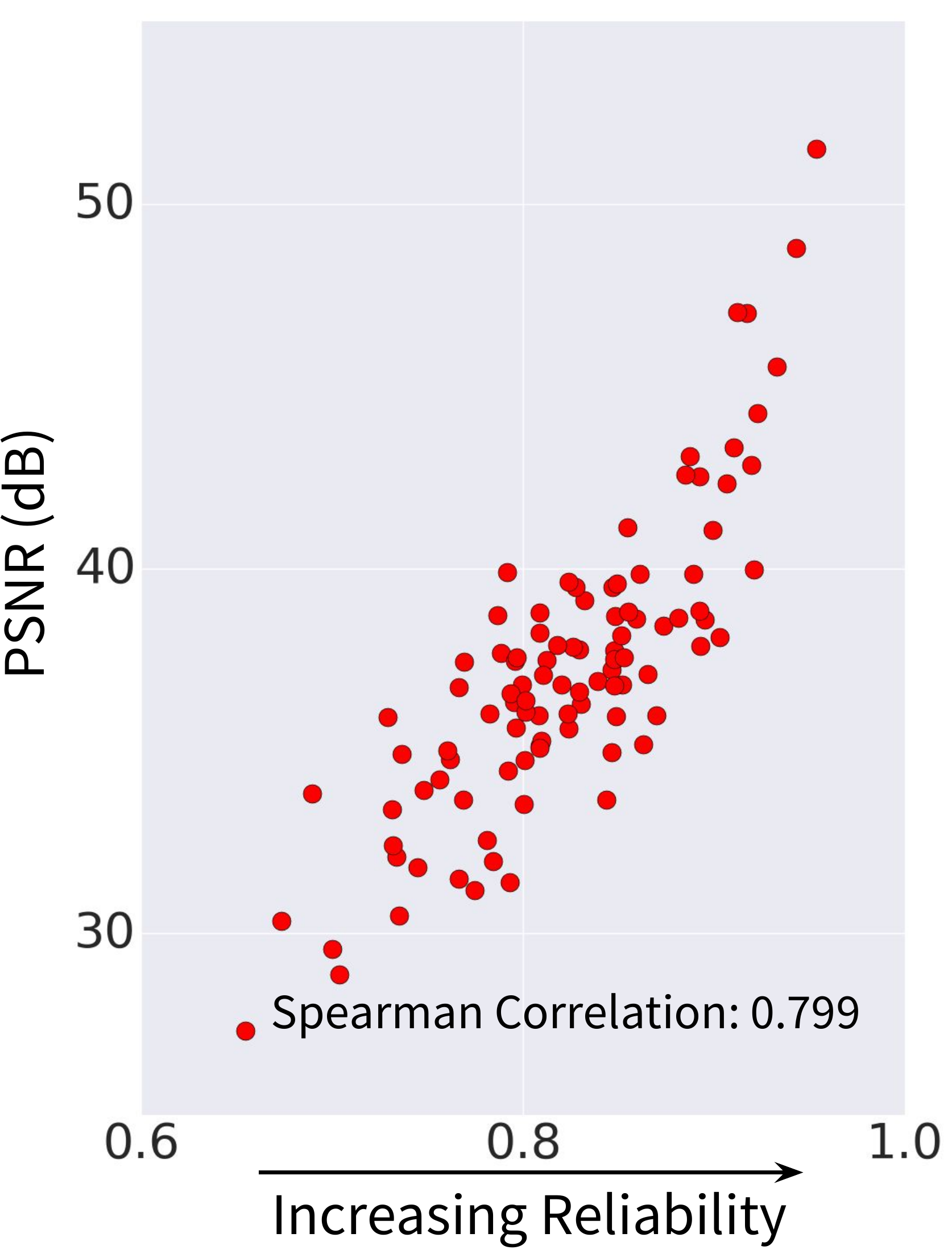}
		\label{fig:psnr_v_conf}
	}\hfill
	\subfigure[Robustness to different views at test time: top left is ground truth, and others are reconstructions using \name\!-adv+WLS.] 
	{\includegraphics[width=0.32 \linewidth,valign=b]{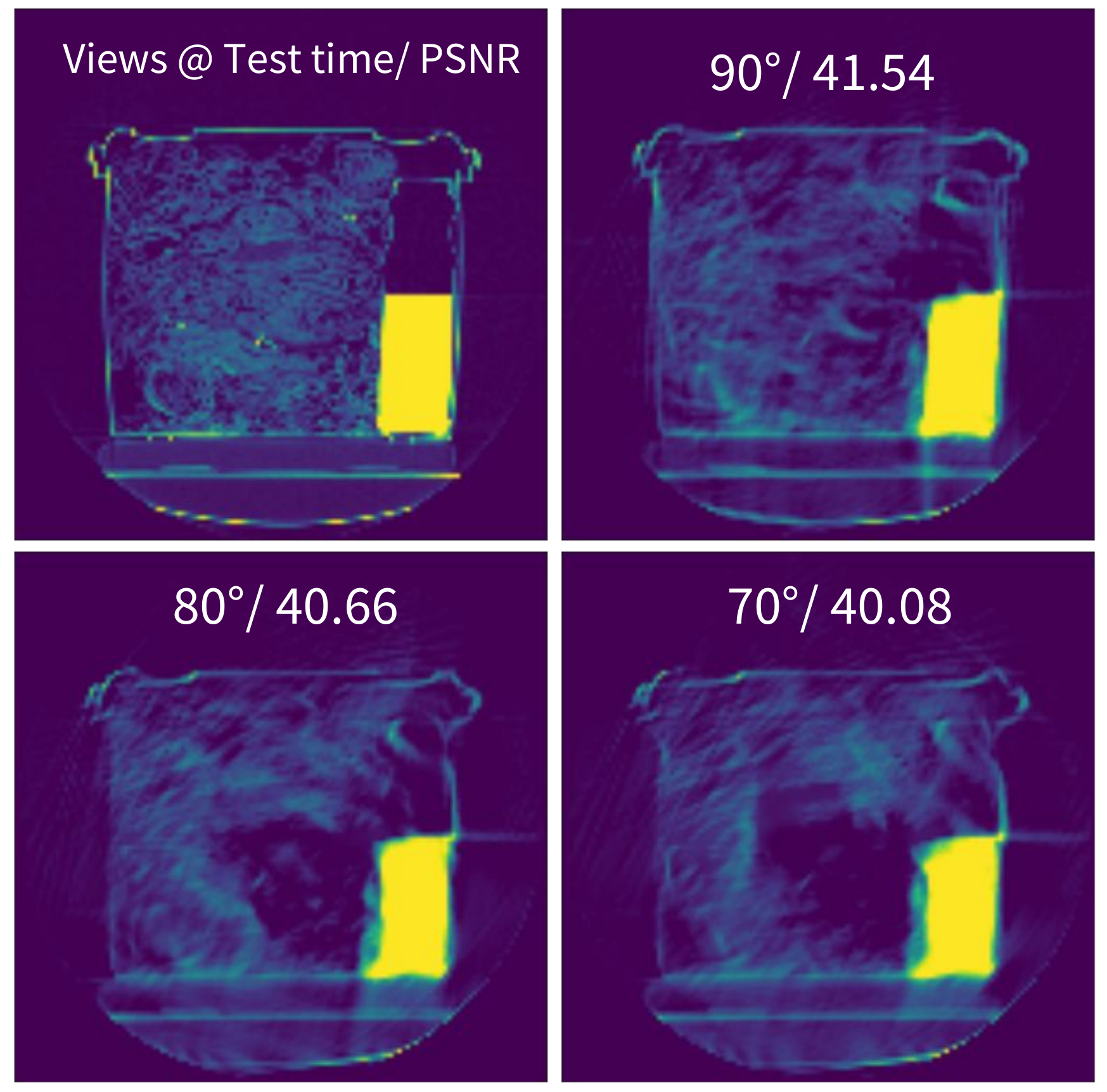}
		\label{fig:robust}
	}

	\caption{\footnotesize{\textbf{Properties of \name\!}: (a) Pixel-wise confidence measures, values are shown on the same scale with values between $(0,3\times 10^{-4})$. (b) Proposed measure of confidence acts as a strong indicator of the quality of reconstruction, (c) Demonstration of the behavior of the proposed approach for varying number of views during testing.}}
	\label{fig:confidence}%
	\vspace{-0.1in}
\end{figure*}

\vspace{5pt}
\noindent \textbf{Setup:} From the training data, we drop rows $(360,720]$ in the sinogram matrix, corresponding to views obtained from $(90^\circ,180^\circ]$. Physically, they represent the viewing angles going 0-90\textdegree clockwise as shown in figure \ref{fig:views}. We consistently drop the same views across all training images and testing images.

\vspace{5pt}
\noindent \textbf{Training details and parameters:} We trained our networks using Tensorflow \cite{abadi2016tensorflow} on a NVIDIA Tesla K40m GPU. We do not perform any scaling on the sinograms or reconstructions, since the value ranges are globally consistent across the entire dataset. Further, we use the same decoder in all our networks, regardless of the loss function. For the $1$D CNN, we used filter sizes $[1,2,3,4,5]$ and employed the Adam optimizer\cite{kingma2014adam} with learning rate $1\times 10^{-3}$ when using the MSE loss. For the adversarial loss, we set the learning rate at $2\times 10^{-4}$ and the exponential decay rate for the first moment estimates at $\beta_1 = 0.5$. 

\subsection{Evaluating Quality of Reconstruction}
The reconstruction results obtained using different baseline solutions and the proposed approach are shown in Figure \ref{fig:ctrecon}. The baseline techniques include the state-of-practice methods, namely FBP and WLS, and variants of \name with both MSE and adversarial losses. We observe that, in general, implicit sinogram completion followed by an analytical reconstruction such as FBP or WLS produces more accurate reconstructions compared to methods that directly predict the CT image (\name\!-mse and \name\!-adv). In particular, \name\!-mse and \name\!-adv performed nearly the same when followed up by WLS, therefore we only show \name\!-adv+WLS in figure \ref{fig:ctrecon} for brevity. For each reconstruction, we compute the PSNR value (dB) and Structural Similarity (SSIM) measures, with respect to the ground truth. From Figure \ref{fig:ctrecon}, it is evident that the proposed solution, in particular with WLS, is significantly better than existing approaches and the baseline architectures. Furthermore, in Table \ref{tab:psnr}, we show the mean PSNR and SSIM for $100$ randomly chosen slices from the test set. Even though our performance is better as measured by PSNR and SSIM in the image space, they are not reflective of the large improvements in the reconstruction quality. Hence, we also measure the PSNR in the sinogram space by forward projecting the images using the X-ray transform, and comparing them to the ground truth $180^\circ$ sinogram. Denoted by S-PSNR in Table \ref{tab:psnr}, the sinogram space PSNR shows that \name is significantly better than existing baseline approaches, and overall the proposed solution of sinogram completion with WLS performs the best in terms of all metrics.

\vspace{5pt}
\noindent \textbf{Confidence Score for Reconstruction:} As described in Section \ref{subsec:confidence}, our proposed confidence score can provide guidance to qualitatively evaluate the reconstructions, without actually generating the ground truth for a test sample. We illustrate the pixel-wise confidence measures in Figure \ref{fig:confidence} for 3 different images, with decreasing levels of reliability obtained using \name\!-adv. Notice that in cases with noisy reconstructions, i.e., cases where the partial views considered do not sufficiently capture the properties of the scene, directly correspond to a lower confidence (or higher variance) as shown by the measure displayed on top of the variance maps. In Figure \ref{fig:psnr_v_conf}, we test the hypothesis that the proposed confidence measure can act as a strong indicator of the actual reconstruction quality, as measured by the PSNR. The strong correlation between PSNR and the proposed metric on the set of test images validates our hypothesis - an overwhelming evidence against the null hypothesis that they are not related, p-value=$1.808 \times 10^{-23}$.

\begin{table}[!htb]
	\caption{\footnotesize{PSNR and SSIM measures comparing 100 randomly sampled test slices with the ground truth. Completing the sinogram with \name followed by WLS is superior to all baseline methods. While PSNR and SSIM are metrics in the image space, S-PSNR is in the sinogram space.}}
	\small{
		\centering
		\begin{tabular}{ |c|c|c|c|}
			\hline
			\cellcolor{gray!30}Method & \cellcolor{gray!30}PSNR (dB) & \cellcolor{gray!30}SSIM & \cellcolor{gray!30}S-PSNR (dB)\\\hline
			FBP & 35.65  &0.846 & 26.56 \\\hline
			WLS & 37.3  & 0.939 & 32.08   \\\hline
			\name-mse 		&  37.70 & 0.940 &  33.70\\\hline
			\name-adv 		& 37.19 &  0.935  & 32.82 \\\hline
			\name-mse + FBP 	& 37.42 &  0.932  & 33.22 \\\hline
			\name-adv + FBP 	& 37.35 &  0.930 & 33.46\\\hline
			\name-mse + WLS & \textbf{38.13} &  \textbf{0.952}  & 34.11 \\\hline
			\name-adv + WLS & {38.08} &  {0.950}  & \textbf{34.14} \\\hline
			
		\end{tabular}
		
		\label{tab:psnr}}
	\vspace{-10pt}
\end{table}

\vspace{5pt}
\noindent \textbf{Testing with fewer views than training:} By design, our formulation allows the use of variable number of views in the sinogram. In practice, this translates to having reasonably stable reconstructions for further reduction in the number of views at test time. Note that, the views that are dropped can be random or can be consecutive in the sequence. Retraining a network for every unique set of views can be an arduous task. Our network can handle these cases well, by producing reconstructions whose quality degrades gracefully with incremental loss in the number of views. An example is shown in figure \ref{fig:robust}, where we show the reconstructions obtained using views ranging from $70^\circ$ to $90^\circ$ during test time.

\begin{figure}[!htb]
	\includegraphics[width=\linewidth]{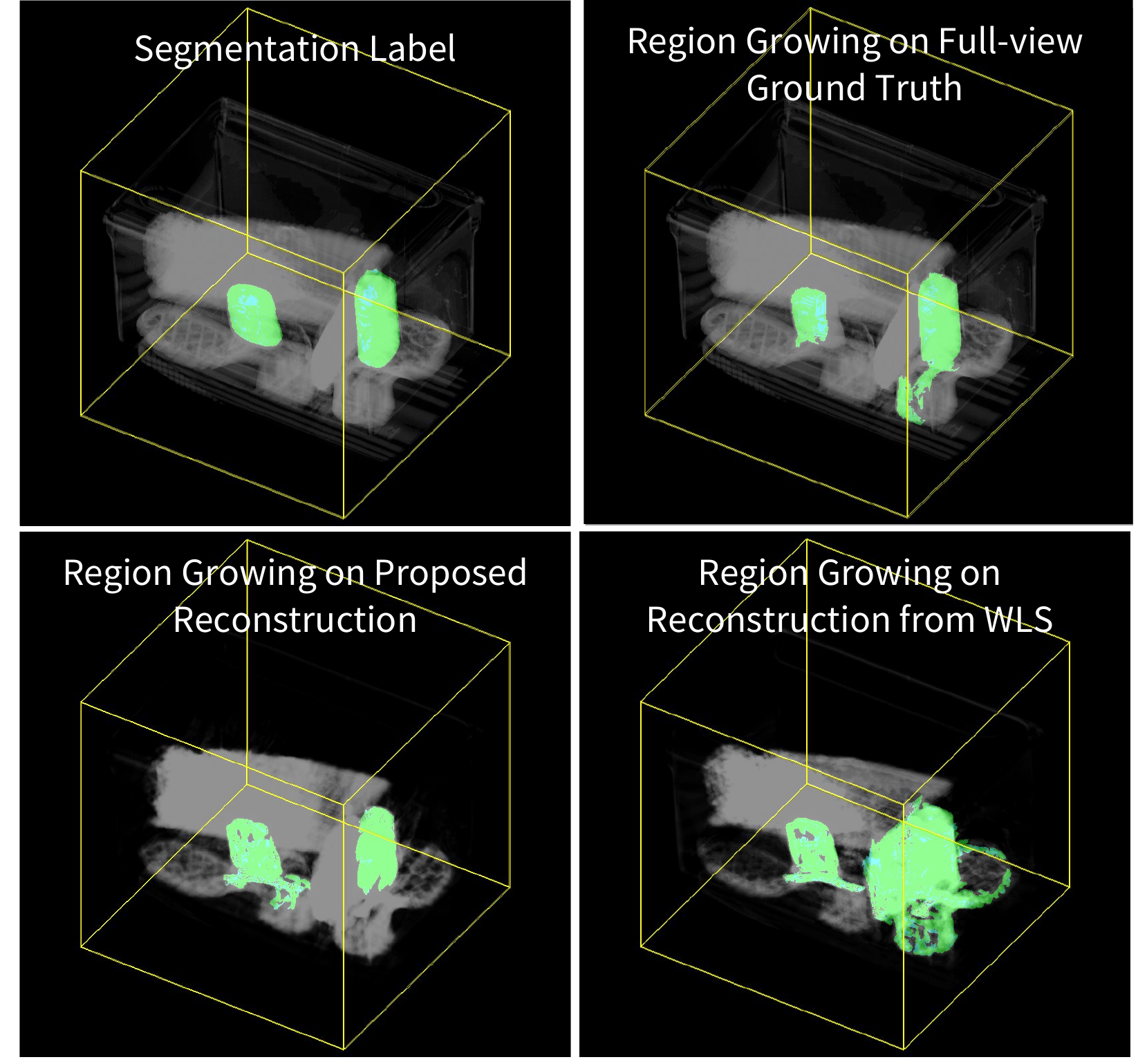}
	\caption{\footnotesize{\textbf{3D Segmentation on limited-view reconstructions:} We employ a region growing 3D segmentation in all cases and the resulting segmentations are shown in color, against a 3D rendering of the reconstructed 2D images underneath. It is clearly evident that our method performs very similar to ground truth in determining the object boundaries compared to WLS. }}
	\label{fig:3Dseg}
	\vspace{-5pt}
\end{figure}

\vspace{5pt}
\noindent \textbf{$3$D Segmentation from CT Reconstructions:} CT images are primarily used to study 3D objects, and hence evaluating the quality of the reconstructions in 3D segmentation can clearly demonstrate their usefulness in practice. We consider the 3D segmentation process, since it is often a critical step prior to performing complex inference tasks such as threat detection \cite{kim2015randomized}. To this end, we use the popular region-growing based segmentation proposed in \cite{wiley2012automatic} to identify high intensity objects in the bags from their reconstructions with partial views. We show an example for a bag with $260$ image slices, that has been rendered in 3D using the 2D slices reconstructed with the proposed \name\!-adv+WLS and WLS alone respectively, in Figure \ref{fig:3Dseg}. We compare the segmentations obtained using our method to the segmentation labels, and those obtained using just WLS. It can be seen in this example, WLS preserves 3D edges poorly resulting in spurious segments, whereas the proposed reconstruction is significantly better, resembling the ground truth. Additional segmentation results can be found in the supplementary material.

\section{Discussion}
In this paper we proposed to accurately recover CT images when the viewing angle is limited to only $90^\circ$. We pose this problem as sinogram completion, but solve it in the image domain. Our empirical studies demonstrate the effectiveness of our three-stage approach -- first computes a neural network based reconstruction, obtains a full-view sinogram using the x-ray transform based on the reconstruction, and then obtains the final reconstruction through WLS on the completed sinogram. We also proposed a confidence score to gauge the reliability of the recovery process, while being a reasonable surrogate for image quality. 
 
\noindent \textbf{Failure Cases:} Our method works best on scenes with large objects that appear as low frequency content in images. We observe no significant gain over WLS in more complicated scenes that contain multiple small objects or intricate designs, since they manifest as high frequency image content, which are very hard to recover in this ill-posed problem setup. However, even with this limitation, we are able to recover 3D structure very well as shown in Figure \ref{fig:3Dseg}, on realistic data that can be of practical use. Finally, since this dataset was not intended for limited angle reconstruction, there are some examples when the objects are just not in view (i.e. completely invisible within 0-90\textdegree), and our network has no information to recover them. 

\noindent \textbf{Future Work:} There are several important directions forward for this work -- (i) include the forward projection step (using x-ray transform) as a final layer inside \name, so that we can optimize the reconstruction setup end-to-end; (ii) Along with the reconstruction, jointly infer the segmentations from the sinograms in a multi-task learning setting. 
\section*{Acknowledgement}
{\small
The first author would also like to thank Kuldeep Kulkarni author of \cite{Reconnet} for several helpful discussions.

\noindent This document was prepared as an account of work sponsored by an agency of the United States government. Neither the United States government nor Lawrence Livermore National Security, LLC, nor any of their employees makes any warranty, expressed or implied, or assumes any legal liability or responsibility for the accuracy, completeness, or usefulness of any information, apparatus, product, or process disclosed, or represents that its use would not infringe privately owned rights. Reference herein to any specific commercial product, process, or service by trade name, trademark, manufacturer, or otherwise does not necessarily constitute or imply its endorsement, recommendation, or favoring by the United States government or Lawrence Livermore National Security, LLC. The views and opinions of authors expressed herein do not necessarily state or reflect those of the United States government or Lawrence Livermore National Security, LLC, and shall not be used for advertising or product endorsement purposes. }

\bibliographystyle{ieee}
\bibliography{refs}
\newpage

\title{\textbf{\Large{SUPPLEMENTARY MATERIAL}}}
\section*{Description of \name}
\begin{figure*}[tb]
\centering
\subfigure[\normalsize{1D CNN}]{
\begin{tabular}{|p{0.17\linewidth}|p{0.3\linewidth}|}
\hline
\textbf{Input}& \textbf{Filters} \\\hline
Sinogram $360\times 128$ & $(f,128,1,256)$, $f = [1,2,3,4,5]$ \\\hline
\end{tabular}
\label{tab:1dcnn}}
\subfigure[\normalsize{Discriminator}]{\begin{tabular}{|p{0.125\linewidth}|p{0.35\linewidth}|}
\hline
\textbf{Input}& \textbf{Filters} \\\hline
I:$128\times 128$ & Conv 2D $(7,7,1,64)\rightarrow \mbox{BN}\rightarrow \mbox{LReLU}$ \\\hline
h1:$64\times 64\times 64$ & Conv 2D $(7,7,64,32)\rightarrow \mbox{BN}\rightarrow \mbox{LReLU}$ \\\hline
h2:$32,768$ & FC $32,768 \times 16\rightarrow \mbox{BN}\rightarrow \mbox{LReLU}$ \\\hline
h3: $16$ & FC $16\times 1 \rightarrow \mbox{BN}\rightarrow \mbox{LReLU}$ \\\hline
\end{tabular}
\label{tab:ctdisc}}

\subfigure[\normalsize{Generator}]{
\includegraphics[width=\linewidth]{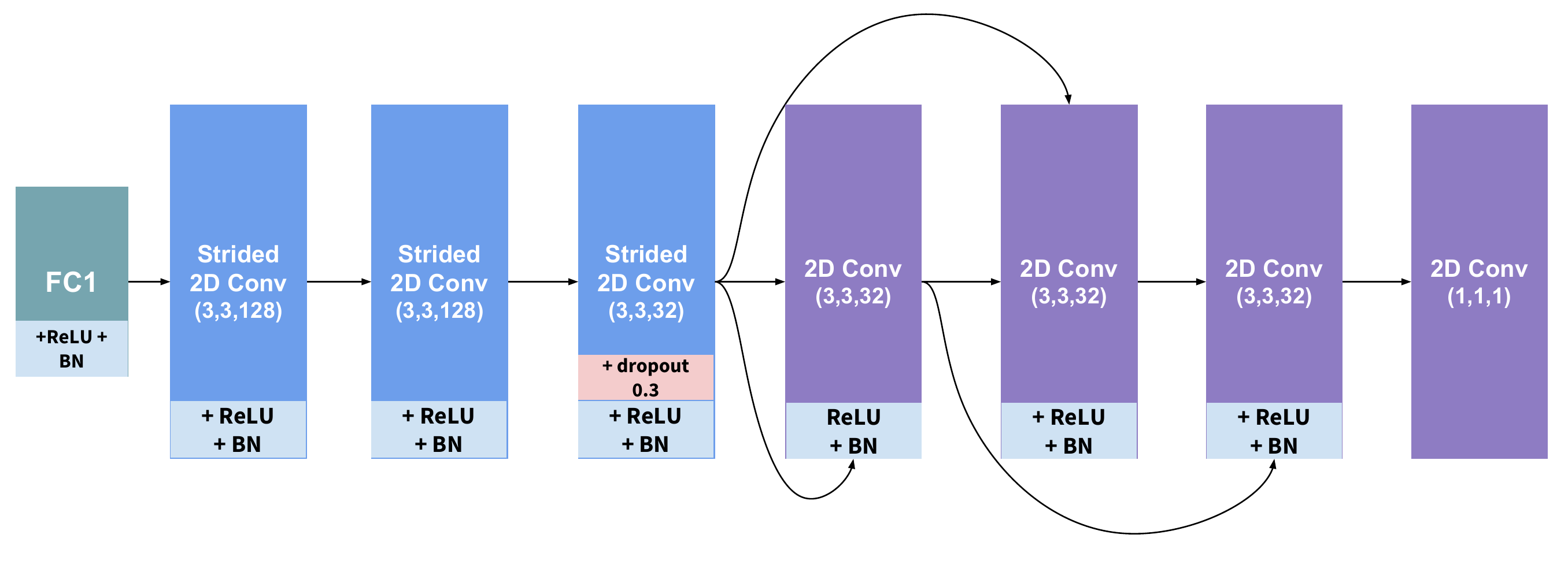}
\label{fig:ctgen}}
\caption{Architecture details of the different networks in CT-Net.}
\end{figure*}

The architectures of the different networks in \name are illustrated in Figure \ref{fig:ctgen}, and tables \ref{tab:1dcnn}, \ref{tab:ctdisc} respectively. We use $5$ different filter sizes, with $256$ hidden dimensions each, for the 1D CNN, which embeds the sinogram into a $5\times 256 = 1280$ dimensional latent representation. Next, we decode the CT image using the decoder described in Figure \ref{fig:ctgen}. Additionally, when using an adversarial loss, we use the discriminator shown in table \ref{tab:ctdisc}.

\begin{figure*}[tb]
\includegraphics[width=\linewidth]{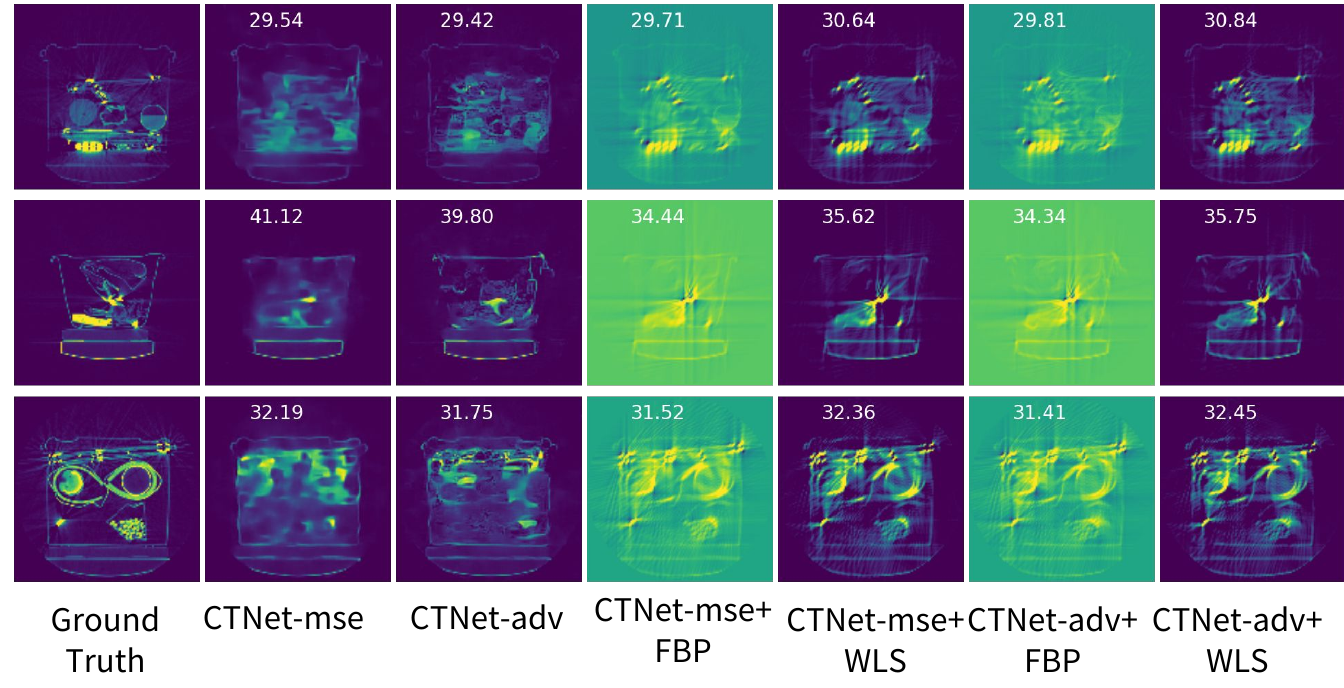}
\caption{\textbf{Failure cases:} Images with high frequency content are much harder to recover, and the proposed method does not provide a significant improvement over existing approaches like FBP or WLS. In addition, if the objects are not visible in the views available, they will naturally appear invisible in the current setup, as seen for the object in the middle row here.}
\label{fig:failure}
\end{figure*}
\section*{Reconstruction Results}
We show failure cases and successful reconstruction examples from unseen test data in Figures \ref{fig:failure} and \ref{fig:ctrecon} respectively. In Figure \ref{fig:ctrecon}, along with the baselines shown in the main paper, we also include \name\!-mse+FBP,\name\!-mse+WLS,\name\!-adv+FBP here for comparison. In general, we observe that sinogram completion works much better than any other approach. Further, sinogram completion with WLS works better than with FBP. Finally, we find \name\!-mse and \name\!-adv perform very similar for completion and do not differ greatly in the final reconstruction. This is because the adversarial loss is measured in the image space. An end-to-end system with an adversarial loss in sinogram completed space is expected to work better, but that is left as future work. From Figure \ref{fig:failure}, it can be observed that test images with lot of high-frequency content are not well recovered with CT-Net. Further, when the objects are not visible in any of the views available, the network has not information to recover that object.

\begin{figure*}[!ht]
\includegraphics[width=\linewidth]{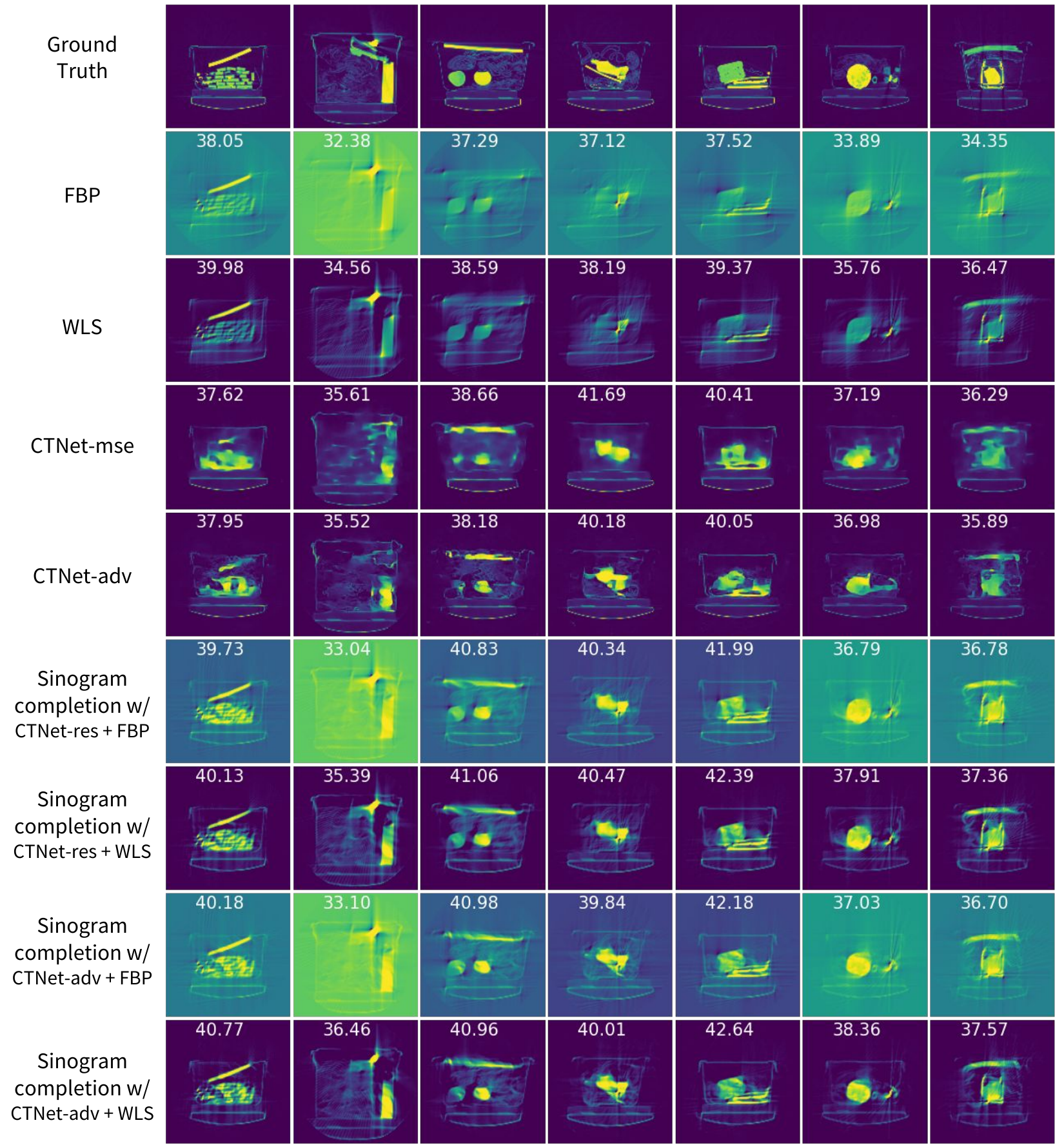}
\caption{\textbf{Successful Cases:} Reconstruction results, with PSNR listed on top of each reconstructed image slice. Apart from the baselines shown in the main paper, we also include \name\!-mse+FBP,\name\!-mse+WLS,\name\!-adv+FBP here for comparison.}
\label{fig:ctrecon}
\end{figure*}

\section*{Generalizability of \name\! to new domains}
\begin{figure*}[!ht]
\subfigure[\normalsize{Comparing training and testing losses for the network warm-started with weights vs randomly initialized. We see that the warm-started model converges within just 50 batches of training, with batch size=100.}]{\includegraphics[width=\linewidth]{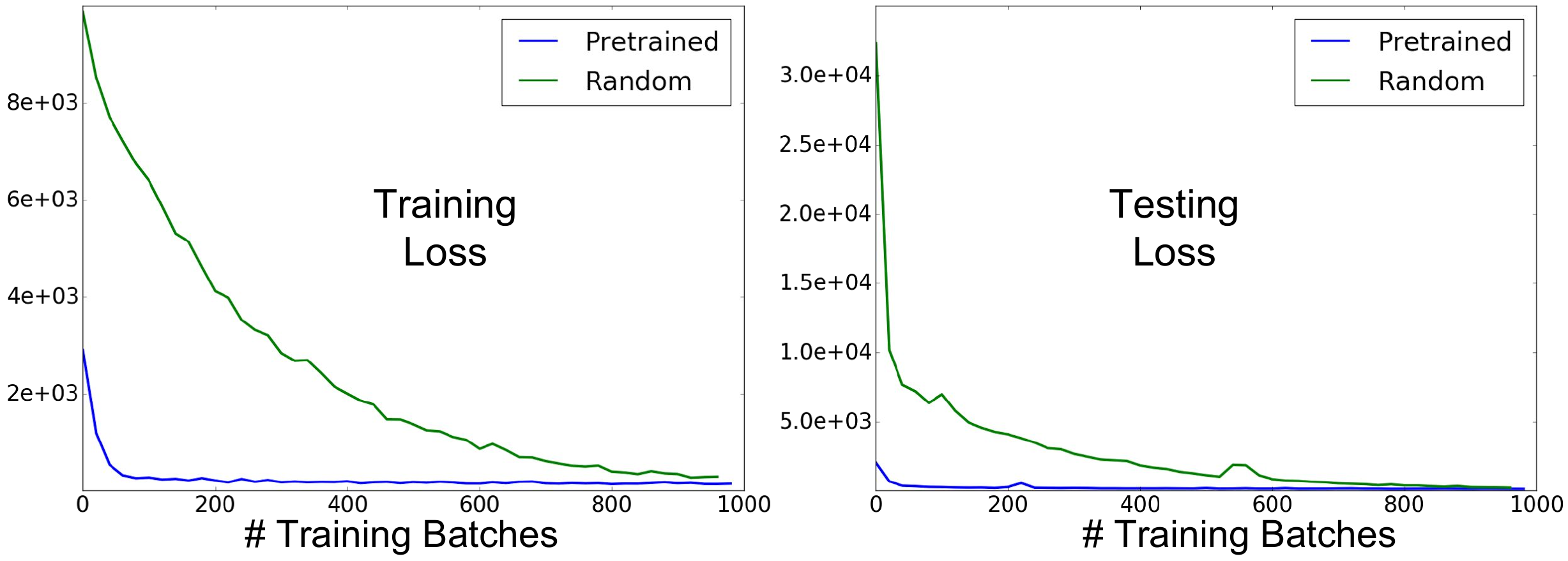} \label{fig:mnist_loss}}

\subfigure[\normalsize{Reconstructed samples from the test set}]{\includegraphics[width=\linewidth]{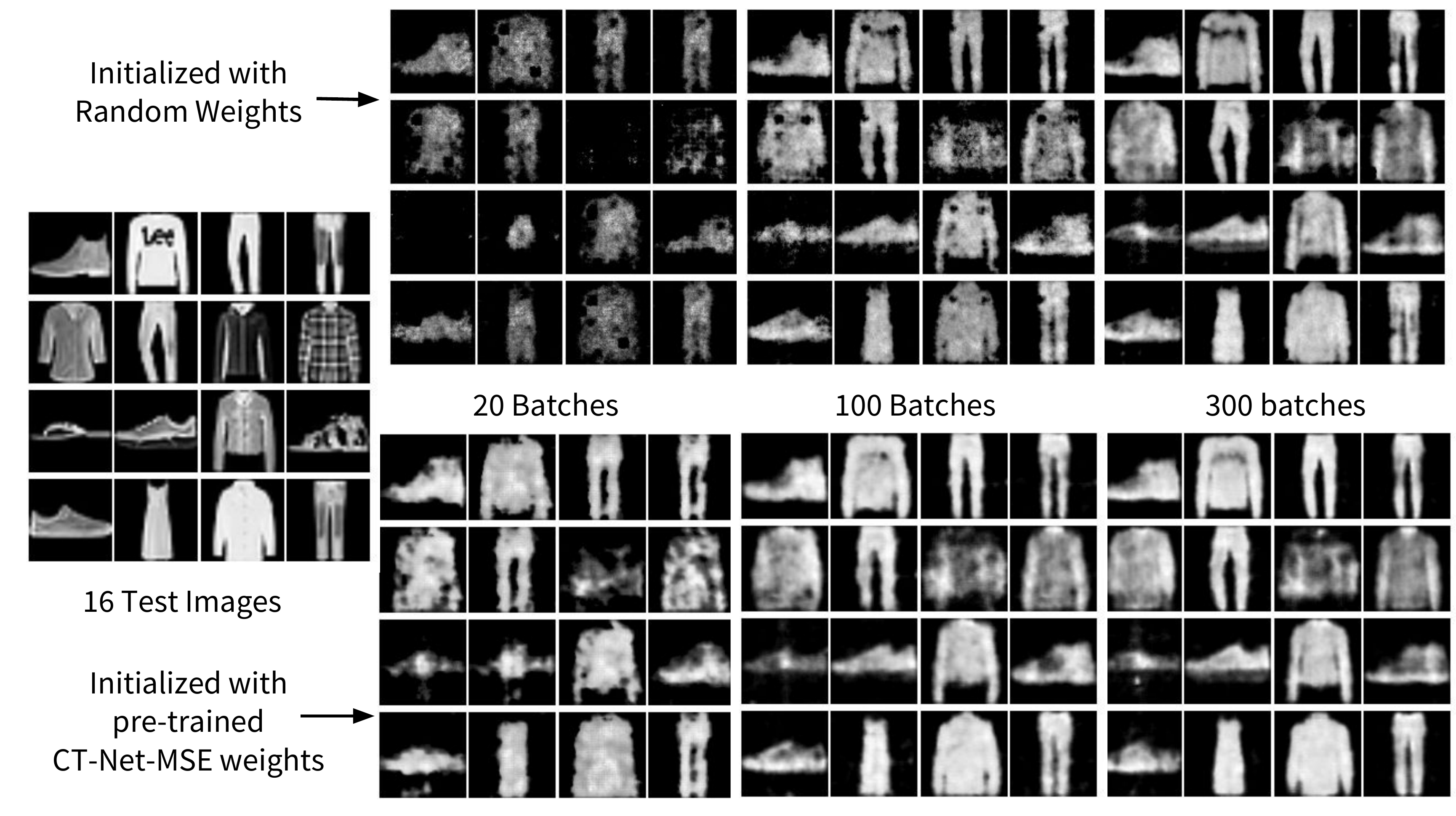} \label{fig:mnist_recon}}
\caption{\textbf{Transferring knowledge to new domains:} We ``warm start'' \name\!, with weights trained on the luggage dataset \cite{coe_data}, and fine-tune for the fashion-MNIST dataset. It can be seen that within just a few batches of training, our network learns to recover the images with just half the views.}
\end{figure*}

A fundamental aspect of our system is its ability to invert the x-ray transform, which maps 3D objects into its corresponding sinogram representation. Conceptually, the inverse transformation should be applicable to any scenario regardless of the objects being scanned. In this experiment, we validate the generalizability of our system by using the network trained on the transportation luggage dataset\cite{coe_data}, to invert the fashion-MNIST \cite{fashion-mnist}. This dataset consists of $60K$ training images of size $28\times 28$ belonging to one of ten classes, consisting of everyday fashion like jeans, dresses, shoes, sandals etc. We resize them to $128\times 128$ using bilinear interpolation. Next, we generate a set of sinograms corresponding to the training data in the fashion-mnist dataset, using the Radon transform which is equivalent to the x-ray transform for 2D images. In particular, we use the implementation of the Radon transform from the scikit-image library \cite{scikit-image}. We fine-tune the pre-trained network from the transportation dataset to the target fashion dataset. For training in the partial view scenario, we use only use the top half of the sinogram, corresponding to $0-90^\circ$. Figure \ref{fig:mnist_loss} shows the training and test losses obtained from training CT-Net on the fashion dataset. As it can be clearly observed, warm-starting the network with weights from the transportation dataset leads to significantly faster convergence within a very few mini-batches, illustrating the effectiveness of the learned inverse x-ray transformation. Furthermore, we illustrate the recovered images in Figure \ref{fig:mnist_recon}.

\section*{3D Segmentation Results}
\begin{figure*}[!htb]
\centering
\subfigure[\normalsize{Test Bag 3 with 270 reconstructed slices.}]{
\includegraphics[width=.65\linewidth]{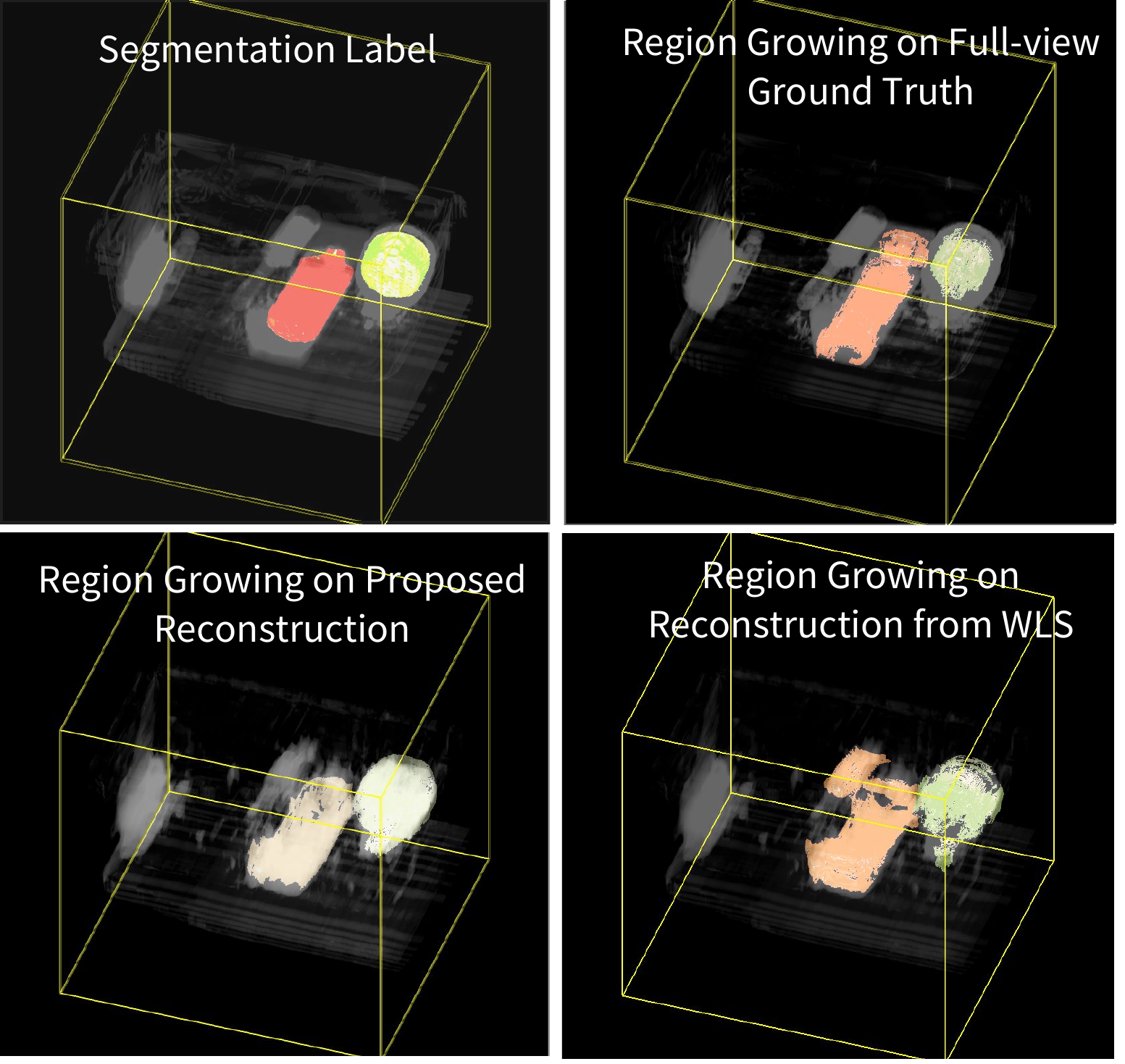}
\label{fig:3dbag2}}

\subfigure[\normalsize{Test Bag 4 with 250 reconstructed slices.}]{
\includegraphics[width=.65\linewidth]{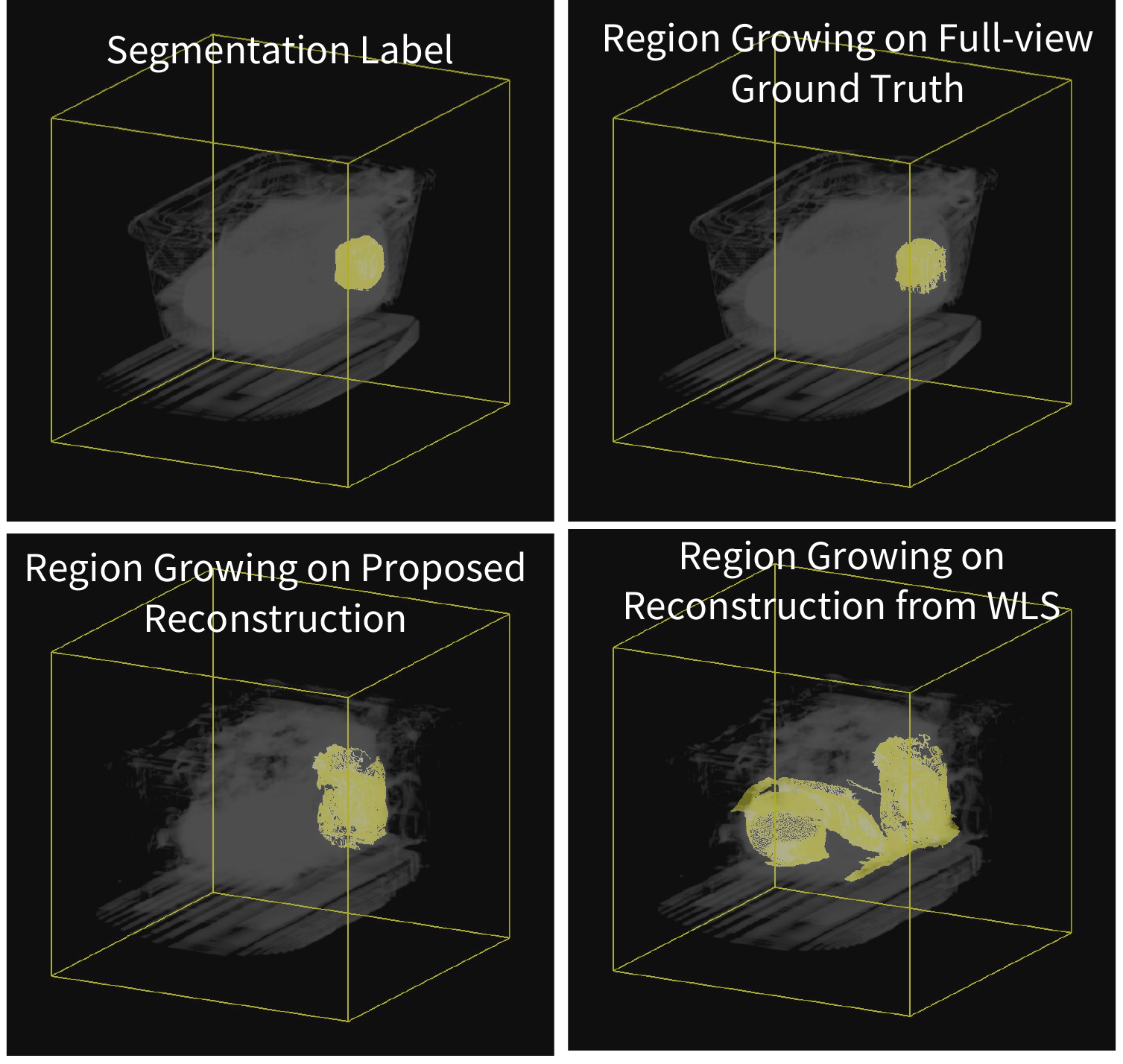}
\label{fig:3dbag3}}
\end{figure*}

\begin{figure*}[!htb]
\centering
\includegraphics[width=.65\linewidth]{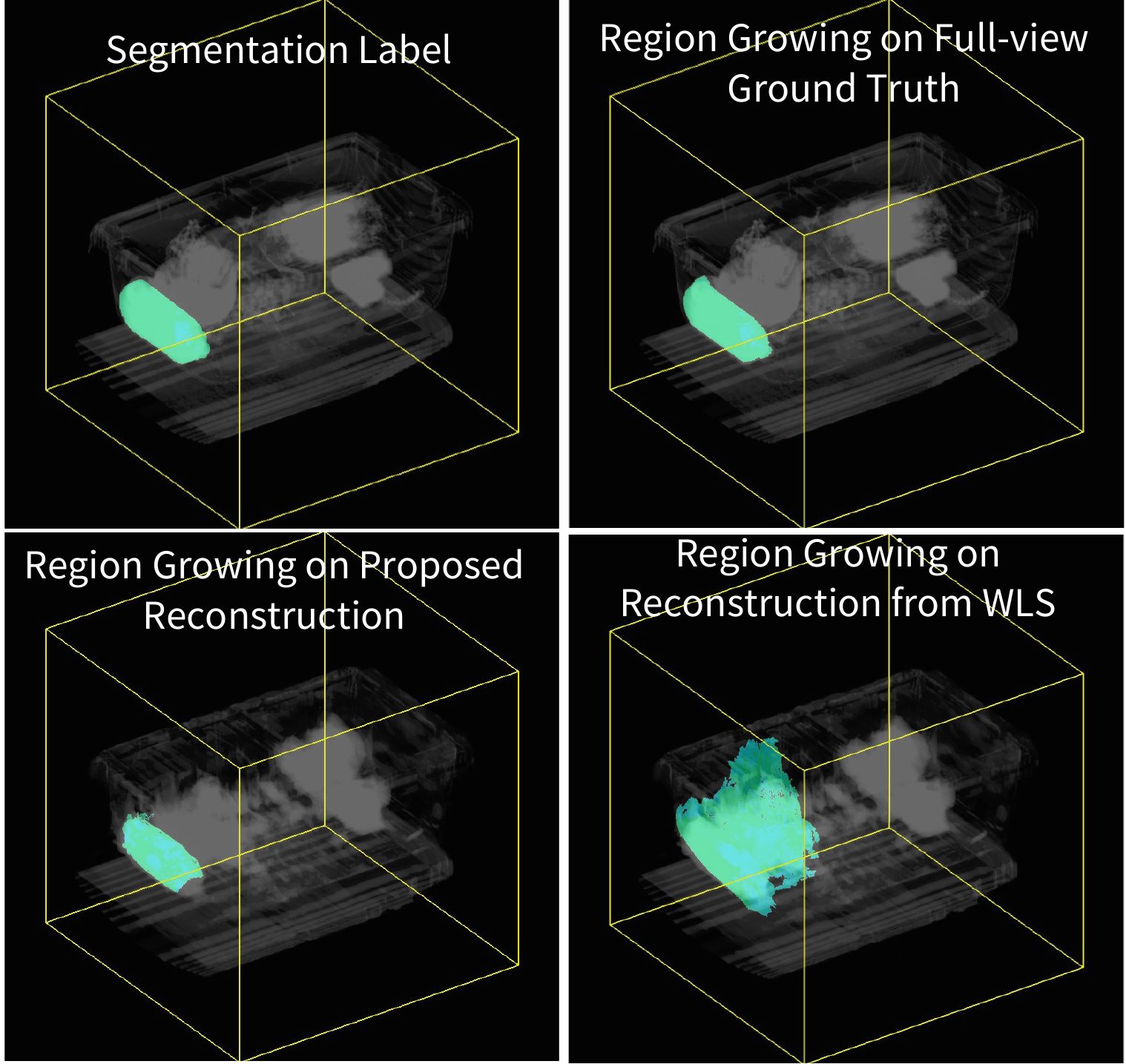}
\caption{Test bag 2 with 274 reconstructed image slices}
\label{fig:3dbag4}
\end{figure*}

In this experiment, we used a popular region growing segmentation similar to the method used in \cite{kim2015randomized}. It is a simplified version of the method in \cite{wiley2012automatic}, with a randomly chosen starting position and a fixed kernel size. The purpose of this experiment is to understand how reconstruction quality affects object segmentation. The luggage dataset contains segmentation labels of objects of interest, and the evaluation focused on how well each segmentation extracts the labeled object. We reconstructed all slices of each bag through the proposed method and combined them into a single bag in 3D. Then we run the region growing in 3D at multiple, hand-tuned parameter settings (intensity threshold ranging from 0.005 to 0.02), and reported the results from the best-performing setting. This is done as some reconstruction results are poor and very sensitive to the threshold (especially when using the partial-view FBP and iterative method). Examples for the 3D segmentation on reconstructions obtained using \name\!-adv+WLS, using region growing method are shown in figures \ref{fig:3dbag2},\ref{fig:3dbag3}, and \ref{fig:3dbag4}. It is easy to see that the proposed reconstruction segments the objects of interest very similar to the ground truth images, than compared to using WLS for reconstruction.


\end{document}